\documentclass{article}

\PassOptionsToPackage{numbers, compress}{natbib}
\usepackage[utf8]{inputenc} 
\usepackage[T1]{fontenc}    
\usepackage{hyperref}       
\usepackage{url}            
\usepackage{booktabs}       
\usepackage{amsfonts}       
\usepackage{nicefrac}       
\usepackage{microtype}      

\usepackage[table]{xcolor}

\newcommand{\be}{\begin{equation}}
\newcommand{\ee}{\end{equation}}
\newcommand{\bel}{\begin{equation}}
\newcommand{\eel}{\end{equation}}
\newcommand{\bea}{\begin{eqnarray}}
\newcommand{\eea}{\end{eqnarray}}
\newcommand{\beal}{\begin{eqnarray}}
\newcommand{\eeal}{\end{eqnarray}}

\usepackage{amsmath}
\usepackage{amssymb}
\usepackage{algorithm,algorithmic}
\usepackage{mathtools}
\usepackage{multirow}

\usepackage{amsthm}

\theoremstyle{plain}
\newtheorem{thm}{Theorem}[section] 

\theoremstyle{definition}
\newtheorem{defn}[thm]{Definition} 
\newtheorem{exmp}[thm]{Example} 
\newtheorem{lemma}[thm]{Lemma}

\usepackage{microtype}
\usepackage{graphicx}
\usepackage{subfigure}
\usepackage{booktabs} 
\usepackage{color}
\usepackage{hyperref}
\usepackage{xr} 

\usepackage{wrapfig}
\usepackage[final]{nips_2018}

\title{A Local Regret in Nonconvex Online Learning}

\author{
	Sergul Aydore \\
	\texttt{Stevens Institute of Technology, NJ, USA}, \\
	\texttt{saydore@stevens.edu} \And
	Lee Dicker \\
	\texttt{Amazon, NY, USA} \\
	\texttt{leehd@amazon.com} \And
	Dean Foster \\
	\texttt{Amazon, NY, USA} \\
	\texttt{foster@amazon.com }
}

\begin{document}
\raggedbottom

\maketitle

\begin{abstract}
We consider an online learning process to forecast a sequence of outcomes for nonconvex models. A typical measure to evaluate online learning algorithms is regret but such standard definition of regret is intractable for nonconvex models even in offline settings. Hence, gradient based definition of regrets are common for both offline and online nonconvex problems. Recently, a notion of local gradient based regret was introduced. Inspired by the concept of calibration and a local gradient based regret, we introduce another definition of regret and we discuss why our definition is more interpretable for forecasting problems. We also provide bound analysis for our regret under certain assumptions.
\end{abstract}

\section{Introduction}
\label{Introduction}
In typical forecasting problems, we make probabilistic estimates of future outcomes based on the previous observations. Recently, it has been shown that forecasting models can be complex nonconvex models \cite{flunkert2017deepar, wen2017multi}. Frequent update of these models is desired as the relationship between the targets and outputs might change over time. However, re-training these models can be time consuming.

Online learning is a method of updating the model on each pattern as it is observed as opposed to batch learning where the training is performed over groups of pattern. It is a common technique to dynamically adapt to new patterns in the data or when training over the entire data set is infeasible. The literature in online learning is rich with interesting theoretical and practical applications but it is usually limited to the convex problems where global optimization is computationally tractable \cite{zinkevich2003online}. On the other hand, it is NP-hard to compute the global minimum of nonconvex functions over a convex domain \cite{hazan2017efficient, hsu2012spectral}.

Due to the intractability of the nonconvex problems, various assumptions on the input have been used to design polynomial-time algorithms \cite{arora2014new, hsu2012spectral}. However, these were too specific to the models and more generic approach was needed. One way to achieve this is by replacing the ``global optimality'' requirement with a more modest requirement of stationarity \cite{allen2016variance}.

The idea of online learning was borrowed from game theory where an online player answers a sequence of questions. The true answers to the questions are unknown to the player at the time of each decision and the player suffers a loss after committing to a decision. These losses are unknown to the player and the performance of the sequence of decisions will be evaluated by the difference between this accumulated loss and the best fixed decision in hindsight. Most recently, \citet{hazan2017efficient} proposed a notion of gradient based local regret for nonconvex games.

Inspired by Hazan's approach and incorporating the notion of calibration, we introduce a novel gradient based local regret for forecasting problems. Calibration is a well-studied concept in forecasting \cite{foster1998asymptotic}. From game theoretic point of view, we call a forecasting procedure ``calibrated'' if the forecasts are consistent in hindsight. To the best of our knowledge, such definition of regret is new. We show that the proposed regret has logarithmic bound under certain circumstances and we provide insights to the proposed regret. We conjecture that more efficient algorithms can be developed that minimizes our regret.

\section{Setting}
In online forecasting, our goal is to update $x_t$ at each $t$ in order to incorporate the most recently available information. Assume that $t \in \mathcal{T} = \left\{ 1, \cdots, T \right\}$ represents a collection of $T$ consecutive points where $T$ is an integer and $t=1$ represents an initial forecast point. $f_1, \cdots, f_T : \mathcal{K} \rightarrow \mathbb{R}$ are nonconvex  loss functions on some convex subset $\mathcal{K} \subseteq \mathbb{R}^d$. To put in another way, $x_t$ represents the parameters of a machine learning model at time $t$, $f_t(x_t)$ represents the loss function computed using the available data at time $t$ given the model parameters $x_t$.

\subsection{Regret Analysis}
The performance of online learning algorithms is commonly evaluated by the regret, which is defined as the difference between the real cumulative loss and the minimum cumulative loss across $\mathcal{T}$:

\be
R(T) \triangleq \sum_{t=1}^T f_t(x_t) - \min_{x \in \mathcal{K}} \sum_{t=1}^T f_t(x).
\ee

If the regret grows linearly with $T$, it can be concluded that the player is not learning. If, on the other hand, the regret grows sub-linearly, the player is learning and its accuracy is improving. While such definition of regret makes sense for convex optimization problems, it is not appropriate for nonconvex problems, due to NP-hardness of nonconvex global optimization even in offline settings. Indeed, most research on nonconvex problems focuses on finding local optima. In literature on nonconvex optimization algorithms, it is common to use the magnitude of the gradient to analyze convergence. \citet{hazan2017efficient} introduced a local regret measure -  a new notion of regret that quantifies the objective of predicting points with small gradients on average. At each round of the game, the gradients of the loss functions from $w$ where $1 \leq w \leq T$ most recent rounds of play are evaluated at the forecast, and these gradients are then averaged. \citet{hazan2017efficient}'s local regret is defined to be the sum of the squared magnitude of the gradients averages.

\begin{defn}
(Hazan's local regret) The $w$-local regret of an online algorithm is defined as:
\be
HR_w(T) \triangleq \sum_{t=1}^T \| \nabla F_{t, w}(x_t) \|^2
\ee
when $\mathcal{K} = \mathbb{R}^d$ and $F_{t,w}(x_t) \triangleq \frac{1}{w} \sum_{i=0}^{w-1} f_{t-i}(x_t)$. \citet{hazan2017efficient} proposed various gradient descent algorithms where the regret $HR$ is sublinear.
\end{defn}
\subsection{Proposed Local Regret}
In order to introduce the concept of calibration \cite{foster1998asymptotic}, let's consider the first order Taylor series expansion of the cumulative loss:
\be
\sum_{t=1}^T f_t(\mbox{proj}_{\mathcal{K}}(x_t + u)) = \sum_{t=1}^T f_t(x_t + D_u(x_t)) \approx \sum_{t=1}^T f_t(x_t) + \sum_{t=1}^T \left\langle D_u(x_t), \nabla f_t(x_t) \right\rangle
\ee
where $D_u(x_t) \triangleq \mbox{proj}_{\mathcal{K}}(x_t + u) - x_t$ for any $u \in \mathbb{R}^d$. If the forecasts $\left\{x_1, \cdots, x_T \right\}$ are well-calibrated, then perturbing $x_t$ by any $u$ cannot substantially reduce the cumulative loss. Hence, we can say that the sequence $\left\{x_1, \cdots, x_T \right\}$ is \textit{asymptotically calibrated} with respect to $\left\{f_1, \cdots, f_T \right\}$, if:
\be
\limsup_{T \rightarrow \infty} \sup_{u \in \mathbb{R}^d} -\frac{1}{T} \sum_{t=1}^T \left\langle D_u(x_t), \nabla f_t(x_t) \right\rangle \leq 0. 
\ee

\begin{defn} (Proposed Regret) We propose a $w$-local regret as:
\be
PR_w(T) \triangleq \sum_{t=1}^T \left \Vert \frac{1}{w} \sum_{s=t-w+1}^t \left \langle  D_u(x_s), \nabla f_s(x_s) \right \rangle \right \Vert^2
\label{eq:proposed_regret}
\ee
\end{defn}
where $f_t(x_t) = 0$ for $t \leq 0$.
To motivate equation \ref{eq:proposed_regret}, we use the following equality:
\be
\lim_{\delta \rightarrow 0} \frac{1}{\delta} \sup_{\|u\|=\delta} \left \Vert \frac{1}{w} \sum_{s=t-w+1}^t \left \langle D_u(x_s), \nabla f_s(x_s) \right \rangle \right \Vert^2 = \left \Vert \frac{1}{w} \sum_{s=t-w+1}^t \nabla f_s(x_s) \right \Vert^2
\ee
which holds for the interior points. Using our definition of regret, we effectively evaluate an online learning algorithm by computing the average of losses at the corresponding forecast values over a sliding window. \citet{hazan2017efficient}'s local regret, on the other hand, computes average of previous losses computed on the most recent forecast. We believe that our definition of regret is more applicable to forecasting problems as evaluating today's forecast on previous loss functions might be misleading.


\section{Bound Analysis}
We provide bound for different scenarios for the proposed regret in equation \ref{eq:proposed_regret} for the interior points in the feasible set with the following assumptions: $\sup_{x, y \in \mathcal{K}} \| x - y\| = M$; $\sup_{x \in \mathcal{K}, t \in \mathcal{T}} \nabla f_t(x) = G$; parameter update at $t$ is: $x_{t+1} = \mbox{proj}_{\mathcal{K}}(x_t - \eta_t \nabla f_t(x_t))$ where $\eta_t = \eta / \sqrt{t}$ is the learning rate for some small $\eta > 0$. We consider three scenarios: (i) $\eta_t = \eta$, $w$ is constant and $\mathcal{K} = \mathbb{R}^d$, (ii) $\eta_t = \eta / \sqrt{t}$ and $w = t$, (iii) $\eta_t = \eta / \sqrt{t}$ and $w$ is constant. We also note the following Theorem whose proof is provided in section \ref{sec:thm_proof}.

\begin{thm}
$\sum_{s = t - w + 1}^t \left\langle D_u(x_s), \nabla f_s(x_s) \right\rangle \geq 2\eta G^2 \sqrt{t-w+1} - \left( \frac{3M^2}{2 \eta} + 2\eta G^2 \right) \sqrt{t}$ where $\eta_s = \eta / \sqrt{s}$.
\label{thm: main}
\end{thm}

\subsection{Scenario 1: $\eta_t = \eta$, $w$ is constant and $\mathcal{K}=\mathbb{R}^d$}
Since $\mathcal{K}=\mathbb{R}^d$, the update rule becomes $x_{t+1} = x_t - \eta \nabla f_t(x_t)$; in other words, no projection operator is necessary. Hence we can write:

\bea
\sum_{s=t-w+1}^t \left\langle D_u(x_s), \nabla f_s(x_s) \right\rangle &=& \sum_{s=t-w+1}^t \left\langle u, \nabla f_s(x_s) \right\rangle = \left\langle u, \sum_{s=t-w+1}^t \nabla f_s(x_s) \right\rangle \\
&=& \left\langle u, \frac{1}{\eta} \sum_{s=t-w+1}^t (x_s - x_{s+1}) \right\rangle = \frac{1}{\eta} \left\langle u, (x_{t-w+1} - x_{t+1}) \right\rangle \nonumber \\
&\leq& \frac{1}{\eta} \| u \| \|x_{t-w+1} - x_{t+1} \| \leq \frac{M \|u\|}{\eta} \nonumber
\eea

Taking $u$ as a unit vector such that $u = \frac{\sum_{s=t-w+1}^t \nabla f_s(x_s)}{\| \sum_{s=t-w+1}^t \nabla f_s(x_s) \|} $,  we can write $\| \sum_{s=t-w+1}^t \nabla f_s(x_s) \|^2 \leq M^2 / \eta^2$. Hence; the bound for the proposed regret becomes: 
\be
PR_w(T) = \sum_{t=1}^T \left \Vert \frac{1}{w} \sum_{s=t-w+1}^t \nabla f_s(x_s) \right \Vert^2 \leq \frac{M^2 T}{w^2 \eta^2}
\ee
which can be made sublinear in $T$ if $w$ is selected large enough.

\subsection{Scenario 2: $\eta_t = \eta / \sqrt{t}$ and $w=t$} \label{sec: scenario2}
Assuming $x_s + u$ is interior of the feasible set for all $u$ and $s$ and setting $w = t$, we can write the result in theorem \ref{thm: main} as:
\bea
\sum_{s=1}^w \left\langle u, \nabla f_s(x_s) \right\rangle = \left\langle u, \sum_{s=1}^t \nabla f_s(x_s) \right\rangle &=& -\frac{ \| \sum_{s=1}^t \nabla f_s(x_s) \|^2 }{\| \sum_{s=1}^t \nabla f_s(x_s) \|} \\
&=&-\| \sum_{s=1}^t \nabla f_s(x_s) \| \\
&\geq& - \left( \frac{3 M^2}{2 \eta} + 2 \eta G^2 \right) \sqrt{t}
\eea
where $u$ is set to $-\frac{  \sum_{s=1}^t \nabla f_s(x_s)  }{\| \sum_{s=1}^t \nabla f_s(x_s) \|}$. Hence, we get:
\bea
\left \| \frac{1}{t} \sum_{s=1}^t \nabla f_s(x_s) \right \|^2 \leq \left( \frac{3 M^2}{2 \eta} + 2 \eta G^2 \right)^2 \frac{1}{t}
\eea
Summing this over $t$ yields:
\bea
PR_w(T) = \sum_{t=1}^T \left \Vert \frac{1}{t} \sum_{s=1}^t \nabla f_s(x_s) \right \Vert^2 \leq \sum_{t=1}^T \left(\frac{3 M^2}{2 \eta} + 2 \eta G^2 \right)^2 \frac{1}{t} \leq \left(\frac{3 M^2}{2 \eta} + 2 \eta G^2 \right)^2 \log(T)
\eea
which concludes the logarithmic bound for the proposed regret for interior points when $\eta_t = \eta / \sqrt{t}$ and $w = t$.
\subsection{Scenario 3: $\eta_t = \eta / \sqrt{t}$ and $w$ is constant}
Similar to \ref{sec: scenario2}, we can write:
\bea
\left \| \frac{1}{w} \sum_{s=1}^t \nabla f_s(x_s) \right \|^2 \leq \left( \frac{3 M^2}{2 \eta} + 2 \eta G^2 \right)^2 \frac{t}{w^2}.
\eea
Summing this result across $t$ yields:
\bea
PR_w(T) = \sum_{t=1}^T \left \| \frac{1}{w} \sum_{s=1}^t \nabla f_s(x_s) \right \|^2 &\leq& \left( \frac{3 M^2}{2 \eta} + 2 \eta G^2 \right)^2 \frac{1}{w^2} \sum_{t=1}^T t \\
&=& \left( \frac{3 M^2}{2 \eta} + 2 \eta G^2 \right)^2 \frac{T (T + 1)}{2 w^2} 
\eea
which is quadratic in $T$ but $w$ can be selected accordingly to make the upper bound sub-linear.

\section{Conclusion}
We introduced a new definition of a local regret to study nonconvex problems in forecasting. We used the concept of a calibration and showed that our regret can be written as a local regret for the interior points in the feasible set. Our regret differs from Hazan's regret in the sense that it emphasizes today's reward as opposed to past reward. We also showed that our definition of regret has a logarithmic bound under some constraints. As a future direction, we plan to study the insights of our regret for the boundary points in the feasible set and propose efficient machine learning algorithms for nonconvex online learning that are optimal in terms of our definition of regret.

\newpage
\bibliographystyle{plainnat}
\bibliography{mybib}

\newpage
\section{Appendix}
\begin{lemma}
$\eta_t \left\langle D_u(x_t), \nabla f_t(x_t) \right\rangle \geq \left\langle u_t - u_{t+1}, u \right\rangle + \frac{1}{2} \left( \|u_{t+1} - x_{t+1} \|^2 - \|u_t - x_t \|^2 \right) - \eta_t^2 G^2$ where $u_t \triangleq \mbox{proj}_{\mathcal{K}}(x_t + u)$, $\sup_{x \in \mathcal{K}, t \in \mathcal{T}} \nabla f_t(x) = G$ for any $u \in \mathcal{K}$ such that $x_t + u \in \mathcal{K}$. 

\begin{proof}
Let $y_{t+1} \triangleq x_t - \eta_t \nabla f_t(x_t)$ and recall that $x_{t+1} = \mbox{proj}_{\mathcal{K}} \left(y_{t+1} \right)$. Then we have:

\bea
\eta_t \left\langle D_u(x_t), \nabla f_t(x_t) \right\rangle &=& \left\langle u_t - x_t, x_t - y_{t+1} \right\rangle = \left\langle u_t - x_t, x_t - x_{t+1} \right\rangle + \left\langle u_t - x_t, x_{t+1} - y_{t+1} \right\rangle \nonumber \\
&=& \left\langle u_t - x_t, x_t - x_{t+1} \right\rangle + \left\langle u_t - x_{t+1}, x_{t+1} - y_{t+1} \right\rangle + \left\langle x_{t+1} - x_t, x_{t+1} - y_{t+1} \nonumber \right\rangle \\
&\geq& \left\langle u_t - x_t, x_t - x_{t+1} \right\rangle + \left\langle x_{t+1} - x_t, x_{t+1} - y_{t+1} \label{eqn:projection} \right\rangle
\eea
The inequality in \ref{eqn:projection} can be justified by geometrical interpretation of projections as shown in Figure \ref{fig:projection_figure}.
%

\begin{figure}[htb!]
  \begin{minipage}{.44\linewidth}
  \vspace*{-2ex}%
\mbox{\hspace*{-.5ex}%
\subfigure[]{\includegraphics[scale=0.24]{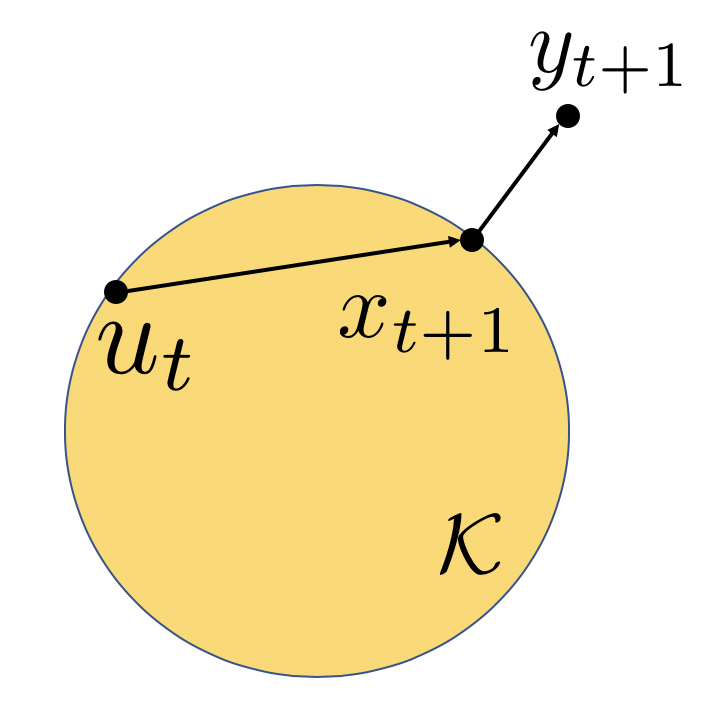}%
    \label{fig:projection_figure}}}%
    \hspace*{-.5ex}%
    \mbox{\subfigure[]{\includegraphics[scale=0.24]{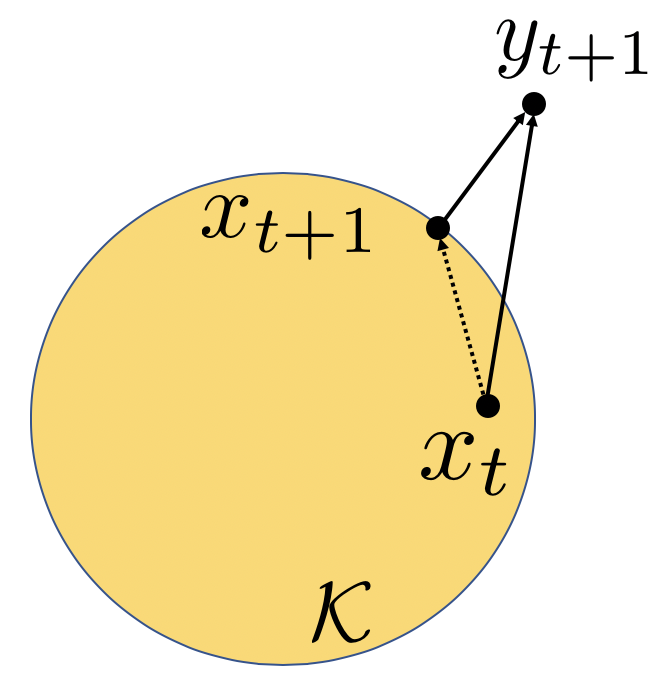} \label{fig:projection_figure_2}}}
\end{minipage}
  \hfill%
\begin{minipage}{.50\linewidth}
\vspace*{-4ex}%
\caption{(a) Geometrical justification for inequality \ref{eqn:projection}. The angle between $u_t - x_{t+1}$ and $x_{t+1} - y_{t+1}$ is always less than or equal to $\pi/2$; hence $\left\langle u_t - x_{t+1}, x_{t+1} - y_{t+1} \right\rangle \geq 0$ for all $u \in \mathbb{R}^d$.
(b) Due to the triangle inequality, $\|x_t - x_{t+1} \| \leq \|x_t - y_{t+1} \| = \eta_t \nabla f_t(x_t)$. Hence $\left\langle x_t - x_{t+1}, \nabla f_t(x_t) \right\rangle \leq \eta_t \| \nabla f_t(x_t) \| ^2 \leq \eta_t G^2$.}
\end{minipage}
\end{figure}

Plugging $y_{t+1} = x_{t+1} - \eta_t \nabla f_t(x_t)$, we have:
\bea
\eta_t \left\langle D_u(x_t), \nabla f_t(x_t) \right\rangle &=& \left\langle u_t - x_t, x_t - x_{t+1} \right\rangle + \left\langle x_t - x_{t+1}, x_t - \eta_t \nabla f_t(x_t) - x_{t+1} \right\rangle \nonumber \\
&=& \left\langle u_t - x_t, x_t - x_{t+1} \right\rangle + \|x_t - x_{t+1} \|^2 - \eta_t \left\langle \nabla f_t(x_t), x_t - x_{t+1} \right\rangle \nonumber \\
&\geq & \left\langle u_t - x_t, x_t - x_{t+1} \right\rangle + \|x_t - x_{t+1} \|^2 - \eta_t^2 G^2 \label{eqn:projection_2}
\eea
Inequality \ref{eqn:projection_2} is a result of triangle inequality as drawn in Figure \ref{fig:projection_figure_2}. Using the fact that $\|u_t - x_t + x_t - x_{t+1} \|^2 = \|u_t - x_t \|^2 + 2 \left\langle u_t - x_t, x_t - x_{t+1} \right\rangle+ \| x_t - x_{t+1} \|^2$ in equation \ref{eqn:projection_2} , we can write:
\bea
\eta_t \left\langle D_u(x_t), \nabla f_t(x_t) \right\rangle &\geq& \frac{1}{2} \left( \|u_t - x_{t+1} \|^2 - \| u_t - x_t \|^2 + \|x_{t+1} - x_t \|^2 \right) - \eta_t^2 G^2 \\
&=& \left\langle u_t - u_{t+1}, u_{t+1} - x_{t+1} \right\rangle + \frac{1}{2} \left( \| u_{t+1} - x_{t+1} \|^2 - \| u_t - x_t \|^2 \right) \nonumber \\
&+& \frac{1}{2} \left( \|u_{t+1} - u_t \|^2 + \| x_{t+1} - x_t \|^2 \right) - \eta_t^2 G^2 \label{eqn:subtract_add}
\eea
where equation \ref{eqn:subtract_add} is a result of $\| u_t - u_{t+1} + u_{t+1} - x_{t+1} \|^2 = \| u_t - u_{t+1} \|^2 + 2 \left\langle u_t - u_{t+1}, u_{t+1} - x_{t+1} \right\rangle + \|u_{t+1} - x_{t+1} \|^2$. By rewriting $\left\langle u_t - u_{t+1}, u_{t+1} - x_{t+1} \right\rangle$ as $\left\langle u_t - u_{t+1}, u_{t+1} - x_{t+1} - u + u \right\rangle$, we get:
\bea
\eta_t \left\langle D_u(x_t), \nabla f_t(x_t) \right\rangle &\geq& \left\langle u_t - u_{t+1}, u \right\rangle + \left\langle u_t - u_{t+1}, u_{t+1} - (x_{t+1} + u) \right\rangle  \nonumber \\
&+& \frac{1}{2} \left( \| u_{t+1} - x_{t+1} \|^2 - \| u_t - x_t \|^2 \right) \nonumber \\
&+& \frac{1}{2} \left( \|u_{t+1} - u_t \|^2 + \| x_{t+1} - x_t \|^2 \right) - \eta_t^2 G^2
\eea
Note that by replacing $x_{t+1}$ with $u_{t+1}$ and $y_{t+1}$ with $x_{t+1} + u$ in Figure \ref{fig:projection_figure}, we can see that $\left\langle u_t - u_{t+1}, u_{t+1} - (x_{t+1} + u) \right\rangle \geq 0$. Since $\frac{1}{2} \left( \|u_{t+1} - u_t \|^2 + \| x_{t+1} - x_t \|^2 \right) \geq 0$, we get:
\be
\eta_t \left\langle D_u(x_t), \nabla f_t(x_t) \right\rangle \geq \left\langle u_t - u_{t+1}, u \right\rangle + \frac{1}{2} \left( \| u_{t+1} - x_{t+1} \|^2 - \| u_t - x_t \|^2 \right) - \eta_t^2 G^2.
\ee
\end{proof}
\label{lemma}
\end{lemma}

\textbf{Proof of Theorem \ref{thm: main}} \label{sec:thm_proof}: \\
As a result of lemma \ref{lemma}, we can write the following inequality:
\bea
\sum_{s=t-w+1}^t \left\langle D_u(x_s), \nabla f_s(x_s) \right\rangle &\geq& \sum_{s=t-w+1}^t \frac{1}{\eta_s} \left\langle u_s - u_{s+1}, u \right\rangle - \sum_{s=t-w+1}^t \eta_s G^2 \nonumber \\ 
&+& \sum_{s=t-w+1}^t \frac{1}{2 \eta_s} \left( \|u_{s+1} - x_{s+1} \|^2 - \| u_s - x_s\|^2 \right) \nonumber
\eea
The first term can be rewritten as
\bea
 \sum_{s=t-w+1}^t \frac{1}{\eta_s} \left\langle u_s - u_{s+1}, u \right\rangle &=& \sum_{s=t-w+1}^t \frac{\sqrt{s}}{\eta} \left\langle u_s - u_{s+1} -x + x, u \right\rangle \\
 &=& \sum_{s=t-w+1}^t \frac{\sqrt{s}}{\eta} \left\langle u_s -x, u \right\rangle - \sum_{s=t-w+1}^t \frac{\sqrt{s}}{\eta} \left\langle u_{s+1} -x, u \right\rangle \nonumber \\
 &=& \frac{t-w+1}{\eta} \left\langle u_{t-w+1} - x, u \right\rangle -\frac{\sqrt{t}}{\eta} \left\langle u_{t+1}-x, u \right\rangle \nonumber \\
 &+& \frac{1}{\eta} \sum_{s=t-w+2}^t \left(\sqrt{s} - \sqrt{s-1} \right) \left\langle u_s - x, u \right\rangle
\eea
The bound for the second term can be written as:
\be
- \eta G^2 \sum_{s=t-w+1}^t \frac{1}{\sqrt{s}} \geq \eta G^2 \left( 2 \sqrt{t-w+1} - 2\sqrt{t} \right) 
\ee
as a result of $\sum_{s=t-w+1}^t \frac{1}{\sqrt{s}} \leq \int_{t-w+1}^t \frac{1}{\sqrt{s}} ds = 2 \sqrt{t} - 2 \sqrt{t-w+1}$.
The bound for the third term can be rewritten as:
\bea
\sum_{s=t-w+1}^t \frac{1}{2 \eta_s} \left( \|u_{s+1} - x_{s+1} \|^2 - \| u_s - x_s\|^2 \right) &=& \sum_{s=t-w+1}^t \frac{\sqrt{s}}{2 \eta} \left( \|u_{s+1} - x_{s+1} \|^2 - \| u_s - x_s\|^2 \right) \nonumber \\
&=& \frac{\sqrt{t}}{2 \eta} \|u_{t+1} - x_{t+1} \|^2 - \frac{\sqrt{t-w+1}}{2 \eta} \|u_{t-w+1} - x_{t-w+1} \|^2 \nonumber \\
&-& \frac{1}{2 \eta} \sum_{s=t-w+2}^t \left(\sqrt{s} - \sqrt{s-1} \right) \|u_s - x_s \|^2 \\
&\geq& \frac{\sqrt{t}}{2 \eta} \|u_{t+1} - x_{t+1} \|^2 - \frac{\sqrt{t-w+1}}{2 \eta} \|u_{t-w+1} - x_{t-w+1} \|^2 \nonumber \\
&-&  \frac{1}{2 \eta} \sum_{s=t-w+2}^t (\sqrt{s} - \sqrt{s-1}) \|u_s - x_s \|^2 \label{eqn:M} \\
&\geq& \frac{\sqrt{t}}{2 \eta} \|u_{t+1} - x_{t+1} \|^2 - \frac{\sqrt{t-w+1}}{2 \eta} \|u_{t-w+1} - x_{t-w+1} \|^2 \nonumber \\
&-&  \frac{M^2}{2 \eta} \underbrace{\sum_{s=t-w+2}^t (\sqrt{s} - \sqrt{s-1})}_{\sqrt{t} - \sqrt{t-w+1}} \\
&=& \frac{\sqrt{t}}{2 \eta} \|u_{t+1} - x_{t+1} \|^2 - \frac{\sqrt{t-w+1}}{2 \eta} \|u_{t-w+1} - x_{t-w+1} \|^2 \nonumber \\
&-& \frac{M^2 \sqrt{t}}{2 \eta} + \frac{M^2 \sqrt{t-w+1}}{2 \eta} \\
&\geq& \frac{\sqrt{t}}{2 \eta} \|u_{t+1} - x_{t+1} \|^2 - \frac{\sqrt{t-w+1}}{2 \eta} M^2 \nonumber \\
&-& \frac{M^2 \sqrt{t}}{2 \eta} + \frac{M^2 \sqrt{t-w+1}}{2 \eta} \\
&=& \frac{\sqrt{t}}{2 \eta} \|u_{t+1} - x_{t+1} \|^2  - \frac{M^2 \sqrt{t}}{2 \eta}  \geq - \frac{M^2 \sqrt{t}}{2 \eta} 
\eea
where equation \ref{eqn:M} is a result of $\sup_{x, y \in \mathcal{K}} \| x - y\| = M$. Hence, we have:
\bea
\sum_{s=t-w+1}^t \left\langle D_u(x_s), \nabla f_s(x_s) \right\rangle &\geq& \frac{t-w+1}{\eta} \left\langle u_{t-w+1} - x, u \right\rangle -\frac{\sqrt{t}}{\eta} \left\langle u_{t+1}-x, u \right\rangle \nonumber \\
 &+& \frac{1}{\eta} \sum_{s=t-w+2}^t \left(\sqrt{s} - \sqrt{s-1} \right) \left\langle u_s - x, u \right\rangle \nonumber \\
&-& \frac{M^2 \sqrt{t}}{2 \eta} + \eta G^2 \left( 2 \sqrt{t-w+1} - 2\sqrt{t} \right) \label{eqn:thm_first_bound} \\
\eea
now, let's explore the bound for $\left\langle u_t - x, u \right\rangle$ for any $x \in \mathcal{K}$. By definition of $u_t$, we can write:
\bea
\|x_t + u - x \|^2 &\geq& \|x_t + u - u_t \|^2 \\
&=& \|x_t + u - x \|^2 + \| x - u_t \|^2 + 2 \left\langle x_t + u - x, x - u_t \right\rangle \\
&=& \|x_t + u - x \|^2 + \| x - u_t \|^2 + 2 \left\langle x_t - x, x - u_t \right\rangle + 2 \left\langle u, x - u_t \right\rangle \\
&\geq&  \|x_t + u - x \|^2 - 2M^2 + 2 \left\langle u, x - u_t \right\rangle.
\label{eqn:another_triangle}
\eea
Hence, $\left\langle u, u_t - x \right\rangle \geq - M^2$. Taking $x = u_{t+1}$ and combining \ref{eqn:thm_first_bound} and \ref{eqn:another_triangle}, we get:
\bea
\sum_{s=t-w+1}^t \left\langle D_u(x_s), \nabla f_s(x_s) \right\rangle &\geq& - \frac{\sqrt{t-w+1}}{\eta} M^2 - \left( \frac{\sqrt{t} - \sqrt{t-w+1}}{\eta} \right)M^2 - \left( \frac{M^2}{2 \eta} + 2 \eta G^2 \right) \sqrt{t}  \nonumber \\
&=& 2 \eta G^2 \sqrt{t-w+1} - \left( \frac{3M^2}{2 \eta} + 2 \eta G^2 \right)\sqrt{t}
\eea

%
%
%
%
%

\end{document}